\documentclass[11pt,a4paper]{article}
\usepackage[hyperref]{emnlp2018}
\usepackage{times}
\usepackage{latexsym}

\usepackage{url}
\usepackage{amsmath}
\usepackage{bm}
\usepackage{booktabs}
\usepackage{multirow}
\usepackage{bm}
\usepackage{dblfloatfix}    
\usepackage{capt-of}
\usepackage{amssymb}
\usepackage{mathabx}
\usepackage [autostyle, english = american]{csquotes}
\usepackage{tabularx}
\usepackage{subcaption}
\usepackage{verbatim}
\usepackage{graphicx}
 \usepackage{lipsum}

\usepackage{algorithm}
\usepackage[noend]{algpseudocode}

\aclfinalcopy 
\def\aclpaperid{1069} 

\newcommand{\code}[1]{\texttt{#1}}
\newcommand{\vt}[1]{\bm{\mathbf{#1}}}

\title{Back-Translation Sampling by Targeting \\Difficult Words in Neural Machine Translation}

\author{Marzieh Fadaee {\normalfont and} Christof Monz\\
Informatics Institute, University of Amsterdam\\
Science Park 904, 1098 XH Amsterdam, The Netherlands\\
  {\tt \{m.fadaee,c.monz\}@uva.nl} \\
 }

\date{}

\begin{document}
\maketitle
\begin{abstract}
Neural Machine Translation has achieved state-of-the-art performance for several language pairs using a combination of parallel and synthetic data.
Synthetic data is often generated by back-translating sentences randomly sampled from monolingual data using a reverse translation model.
While back-translation has been shown to be very effective in many cases, it is not entirely clear why.
In this work, we explore different aspects of back-translation, and show that words with high prediction loss during training benefit most from the addition of synthetic data.
We introduce several variations of sampling strategies targeting difficult-to-predict words using prediction losses and frequencies of words. In addition, we also target the contexts of difficult words and sample sentences that are similar in context.   
Experimental results for the WMT news translation task show that our method improves translation quality by up to 1.7 and 1.2 \textsc{Bleu} points over back-translation using random sampling for German$\rightarrow$English and English$\rightarrow$German, respectively.
\end{abstract}

\section{Introduction}

Neural machine translation (NMT) using a sequence-to-sequence model has achieved state-of-the-art performance for several language pairs \cite{DBLP:journals/corr/BahdanauCB14,sutskever2014sequence,cho2014properties}.
The availability of large-scale training data for these sequence-to-sequence models is essential for achieving good translation quality.

Previous approaches have focused on leveraging monolingual data which is available in much larger quantities than parallel data \cite{Lambert:2011:ITM:2132960.2132997}. \newcite{gulcehre2017integrating} proposed two methods, shallow and deep fusion, for integrating a neural language model into the NMT system.
They observe improvements by combining the scores of a neural language model trained on target monolingual data with the NMT system.

\newcite{sennrich-haddow-birch:2016:P16-11} proposed back-translation of monolingual target sentences to the source language and adding the synthetic sentences to the parallel data.
In this approach a reverse model trained on parallel data is used to translate sentences from target-side monolingual data into the source language. 
This \textit{synthetic} parallel data is then used in combination with the actual parallel data to re-train the model.
This approach yields state-of-the-art results even when large parallel data are available and has become common practice in NMT \cite{2017arXiv170800726S,2017arXiv170704499G,2017arXiv171107893H}. 

While back-translation has been shown to be very effective to improve translation quality, it is not exactly clear why it helps. Generally speaking, it mitigates the problem of overfitting and fluency by exploiting additional data in the target language. 
An important question in this context is how to select the monolingual data in the target language that is to be back-translated into the source language to optimally benefit translation quality. \newcite{pham2017karlsruhe} experimented with using domain adaptation methods to select monolingual data based on the cross-entropy between the monolingual data and in-domain corpus \cite{axelrod2015class} but did not find any improvements over random sampling as originally proposed by \newcite{sennrich-haddow-birch:2016:P16-11}.
Earlier work has explored to what extent data selection of parallel corpora can benefit translation quality \cite{D11-1033,vanderwees-bisazza-monz:2017:EMNLP2017}, but such selection techniques have not been investigated in the context of back-translation. 

In this work, we explore different aspects of the back-translation method to gain a better understanding of its performance.
Our analyses show that the quality of the synthetic data acquired with a reasonably good model has a small impact on the effectiveness of back-translation,
but the ratio of synthetic to real training data plays a more important role.
With a higher ratio, the model gets biased towards noises in synthetic data and unlearns the parameters completely.
Our findings show that it is mostly words that are difficult to predict in the target language that benefit from additional back-translated data. 
These are the words with high prediction loss during training when the translation model converges.
We further investigate these difficult words and explore alternatives to random sampling of sentences with a focus on increasing occurrences of such words.

Our proposed approach is twofold: identifying difficult words and sampling with the objective of increasing occurrences of these words, 
and identifying contexts where these words are difficult to predict and sample sentences similar to the difficult contexts.
With targeted sampling of sentences for back-translation we achieve improvements of up to 1.7 \textsc{Bleu} points over back-translation using random sampling.

\section{Back-Translation for NMT}

In this section, we briefly review a sequence-to-sequence NMT system and describe our experimental settings.
We then investigate different aspects and modeling challenges of integrating the back-translation method into the NMT pipeline.

\subsection{Neural Machine Translation}\label{nmtsec}

The NMT system used for our experiments is an encoder-decoder network with recurrent architecture \cite{luong:2015:EMNLP}.
For training the NMT system, two sequences of tokens, $X =  \big[ x_1, \ldots, x_n \big] $ and $Y =  \big[ y_1, \ldots, y_m \big] $, are given in the source and target language, respectively.

The source sequence is the input to the encoder which is a bidirectional long short-term memory network generating a representation $\vt{s}_n$.
Using an attention mechanism \cite{DBLP:journals/corr/BahdanauCB14}, the attentional hidden state is:
\begin{align*}
\widetilde{\vt{h}}_t = \tanh(\vt{W}_c[\vt{c}_t; \vt{h}_t])
\end{align*}
where $\vt{h}_t$ is the target hidden state at time step $t$ and $\vt{c}_t$ is the context vector which is a weighted average of $\vt{s}_n$.

The decoder predicts each target token $y_t$ by computing the probability:
\begin{align*}
p(y_t | y_{<t}, \vt{s}_n) = \textup{softmax}(\vt{W}_o\widetilde{\vt{h}}_t)
\end{align*}

For the token $y_t$, the conditional probability $p(y_t | y_{<t}, \vt{s}_n)$ during training quantifies the difficulty of predicting that token in the context $y_{<t}$.
The prediction loss of token $y_t$ is the negative log-likelihood of this probability.

During training on a parallel corpus $\mathbb{D}$, the cross-entropy objective function is defined as:
\begin{align*}
\mathcal{L} = \sum_{(X,Y) \in \mathbb{D}} \sum_{i=1}^{m} - \log p(y_i | y_{<i}, \vt{s}_n)
\end{align*}

The objective of this function is to improve the model's estimation of predicting target words given the source sentence and the target context. 

The model is trained end-to-end by minimizing the negative log likelihood of the target words.

\subsection{Experimental Setup}

For the translation experiments, we use English$\leftrightarrow$German WMT17 training data and report results on newstest 2014, 2015, 2016, and 2017 \cite{bojar-EtAl:2017:WMT1}. 

As NMT system, we use a 2-layer attention-based encoder-decoder model implemented in OpenNMT \cite{2017opennmt} trained with embedding size 512, hidden dimension size 1024, and batch size 64.
We pre-process the training data with Byte-Pair Encoding (BPE) using 32K merge operations \cite{sennrich-haddow-birch:2016:P16-12}. 

We compare the results to~\newcite{sennrich-haddow-birch:2016:P16-11} by back-translating random sentences from the monolingual data and combine them with the parallel training data.
We perform the random selection and re-training 3 times and report the averaged outcomes for the 3 models.
In all experiments the sentence pairs are shuffled before each epoch.

We measure translation quality by single-reference case-sensitive \textsc{Bleu} \cite{Papineni2001} computed with the \code{multi-bleu.perl} script from Moses.

\subsection{Size of the Synthetic Data in Back-Translation}

One selection criterion for using back-translation is the ratio of real to synthetic data.
\newcite{sennrich-haddow-birch:2016:P16-11} showed that higher ratios of synthetic data leads to decreases in translation performance. 

\setlength{\tabcolsep}{4pt}
\begin{table}[htb!]
\begin{center}
\small
\begin{tabular}{lrrrrr}
 \toprule \bf  & \bf 	Size &  \bf 2014 &  \bf	2015 &  \bf	2016 &  \bf	2017 \\ \midrule%
 Baseline	& 4.5M	& 26.7	&27.6	&32.5	&28.1\\
\hspace{.15cm} + synthetic (\mbox{1:1})	& 9M	&	28.7&	29.7	&36.3	&30.8\\
\hspace{.15cm} + synthetic (\mbox{1:4}) 	&23M	&	29.1&	30.0&	36.9	&31.1\\
\hspace{.15cm} + synthetic (\mbox{1:10})  & 50M	&22.8&23.6&	29.2	&		 23.9\\
\bottomrule
\end{tabular}
\end{center}
\caption{\label{bigbigger} German$\rightarrow$English translation quality (\textsc{Bleu}) of systems with different ratios of \textit{\mbox{real:syn}} data.}
\end{table}

\begin{figure}[htb]
\includegraphics[width=0.5\textwidth]{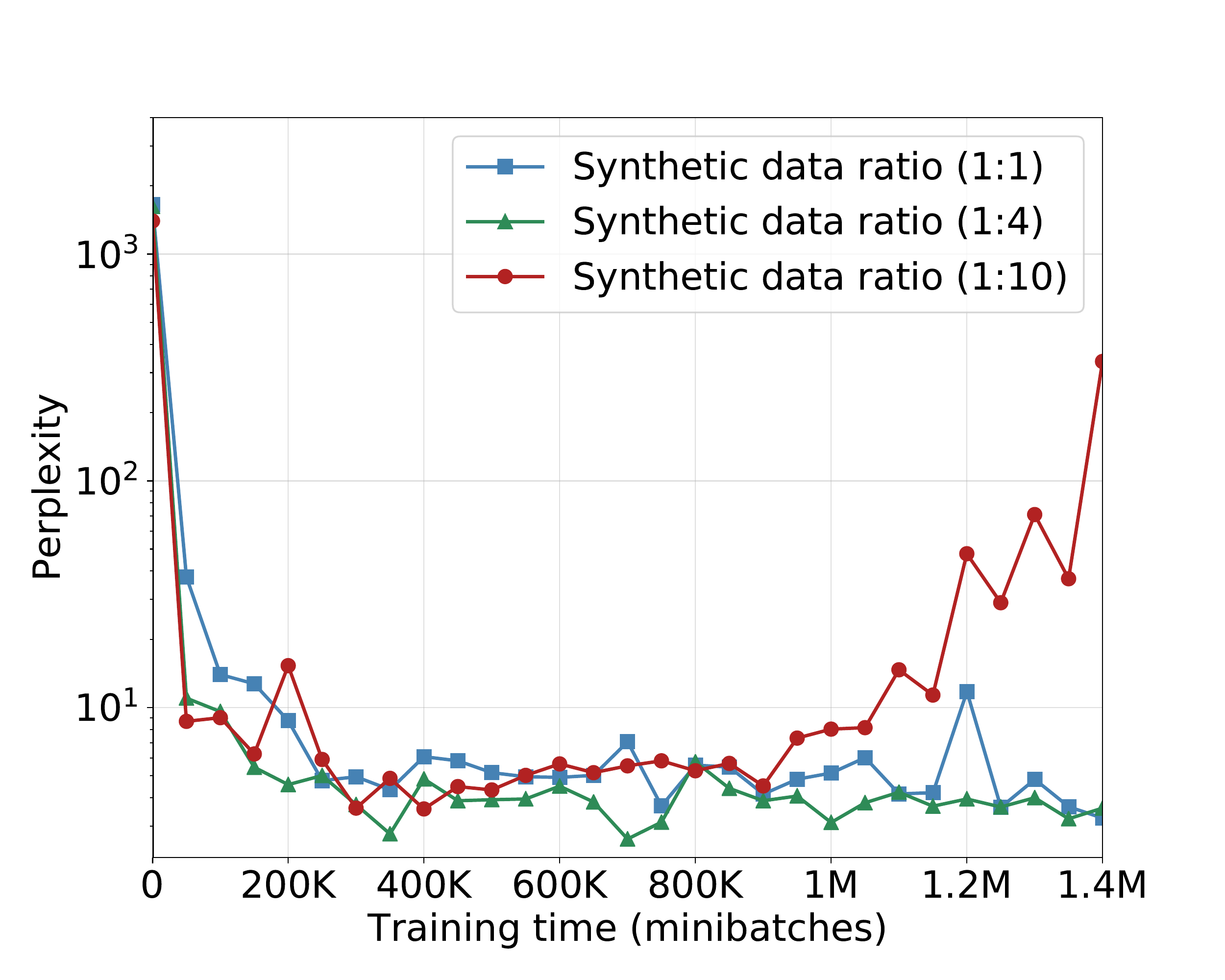}
\caption{Training plots for systems with different ratios of  (\mbox{$real:syn$}) training data, showing perplexity on development set.}
\label{ppl}
\end{figure}

In order to investigate whether the improvements in translation performance increases with higher ratios of synthetic data, we perform three experiments with different sizes of synthetic data.
Figure~\ref{ppl} shows the perplexity as a function of training time for different sizes of synthetic data.
One can see that all systems perform similarly in the beginning and converge after observing increasingly more training instances.
However, the model with the ratio of (\mbox{1:10}) synthetic data gets increasingly biased towards the noisy data after 1M instances.
Decreases in performance with more synthetic than real data is also inline with findings of \newcite{2018arXiv180406189P}.

Comparing the systems using ratios of (\mbox{1:1}) and (\mbox{1:4}), we see that the latter achieves lower perplexity during training.
Table~\ref{bigbigger} presents the performance of these systems on the German$\rightarrow$English translation task. 
The \textsc{Bleu} scores show that the translation quality does not improve linearly with the size of the synthetic data.
The model trained on (\mbox{1:4}) real to synthetic ratio of training data achieves the best results, however, the performance is close to the model trained on (\mbox{1:1}) training data.

\subsection{Direction of Back-Translation}

Adding monolingual data in the target language to the training data has the benefit of introducing new context and improving the fluency of the translation model.
The automatically generated translations in the source language while being erroneous, introduce new context for the source words and will not affect the translation model significantly.

Monolingual data is available in large quantities for many languages. 
The decision of the direction of back-translation is subsequently not based on the monolingual data available, but on the advantage of having more fluent source or target sentences.

\newcite{Lambert:2011:ITM:2132960.2132997} show that adding synthetic source and real target data achieves improvements in traditional phrase-based machine translation (PBMT). Similarly in previous works in NMT, back-translation is performed on monolingual data in the target language.

We perform a small experiment to measure whether back-translating from source to target is also beneficial for improving translation quality.
Table~\ref{dir} shows that in both directions the performance of the translation system improves over the baseline. 
This is in contrast to the findings of~\newcite{Lambert:2011:ITM:2132960.2132997} for PBMT systems where they show that using synthetic target data does not lead to improvements in translation quality.

Still, when adding monolingual data in the target language the \textsc{Bleu} scores are slightly higher than when using monolingual data in the source language. This indicates the moderate importance of having fluent sentences in the target language.

\begin{table}[t!]
\small
\begin{center}
\begin{tabular}{lrrrrr}
 \toprule \bf  & \bf 	Size &  \bf 2014 &  \bf	2015 &  \bf	2016 &  \bf	2017 \\ \midrule%
Baseline	& 4.5M	&	21.2&	23.3&	28.0&	22.4\\
\hspace{.15cm} + synthetic tgt &	9M		&22.4	&25.3	&29.8	&23.7\\
\hspace{.15cm} + synthetic src &	9M		&24.0&	26.0	&30.7	&24.8\\
\bottomrule
\end{tabular}
\end{center}
\caption{\label{dir} English$\rightarrow$German translation quality (\textsc{Bleu}) of systems using forward and reverse models for generating synthetic data.}
\end{table}

\subsection{Quality of the Synthetic Data in Back-Translation}

One selection criterion for back-translation is the quality of the synthetic data.
\newcite{W18-2709} studied the effects of noise in the training data on a translation model and discovered that NMT models are less robust to many types of noise than PBMT models.
In order for the NMT model to learn from the parallel data, the data should be fluent and close to the manually generated translations.
However, automatically generating sentences using back-translation is not as accurate as manual translations.

\begin{table}[htb!]
\begin{center}
\small
\begin{tabular}{lrrrrr}
 \toprule \bf  & \bf  Size  &  \bf 2014 &  \bf	2015 &  \bf	2016 &  \bf	2017 \\ \midrule%
Baseline  &	2.25M	&24.3	&24.9	&29.5	&25.6\\
\hspace{.15cm} + synthetic  &	4.5M	 &26.0	&26.9	&32.2	&27.5\\
\hspace{.15cm} + ground truth	& 4.5M	& 26.7	&27.6	&32.5	&28.1\\
\bottomrule
\end{tabular}
\end{center}
\caption{\label{groundt} German$\rightarrow$English translation quality (\textsc{Bleu}). }
\end{table}

To investigate the \textit{oracle gap} between the performance of manually generated and back-translated sentences, we perform a simple experiment using the existing parallel training data.
In this experiment, we divide the parallel data into two parts, train the reverse model on the first half of the data and use this model to back-translate the second half.
The manually translated sentences of the second half are considered as ground truth for the synthetic data.

Table~\ref{groundt} shows the \textsc{Bleu} scores of the experiments.
As to be expected, re-training with additional parallel data yields higher performance than re-training with additional synthetic data.
However, the differences between the \textsc{Bleu} scores of these two models are surprisingly small.
This indicates that performing back-translation with a reasonably good reverse model already achieves results that are close to a system that uses additional manually translated data.
This is inline with findings of \newcite{sennrich-haddow-birch:2016:P16-11} who observed that the same monolingual data translated with three translation systems of different quality and used in re-training the translation model yields similar results.

\section{Back-Translation and Token Prediction}

In the previous section, we observed that the reverse model used to back-translate achieves results comparable to manually translated sentences.
Also, there is a limit in learning from synthetic data, and with more synthetic data the model unlearns its parameters completely. 

In this section, we investigate the influence of the sampled sentences on the learning model.
\newcite{fadaee-bisazza-monz:2017:Short2} showed that 
targeting specific words during data augmentation improves the generation of these words in the right context. 
Specifically, adding synthetic data to the training data has an impact on the prediction probabilities of individual words.
In this section, we further examine the effects of the back-translated synthetic data on the prediction of target tokens.

\begin{figure}[htb!]
\includegraphics[width=0.5\textwidth]{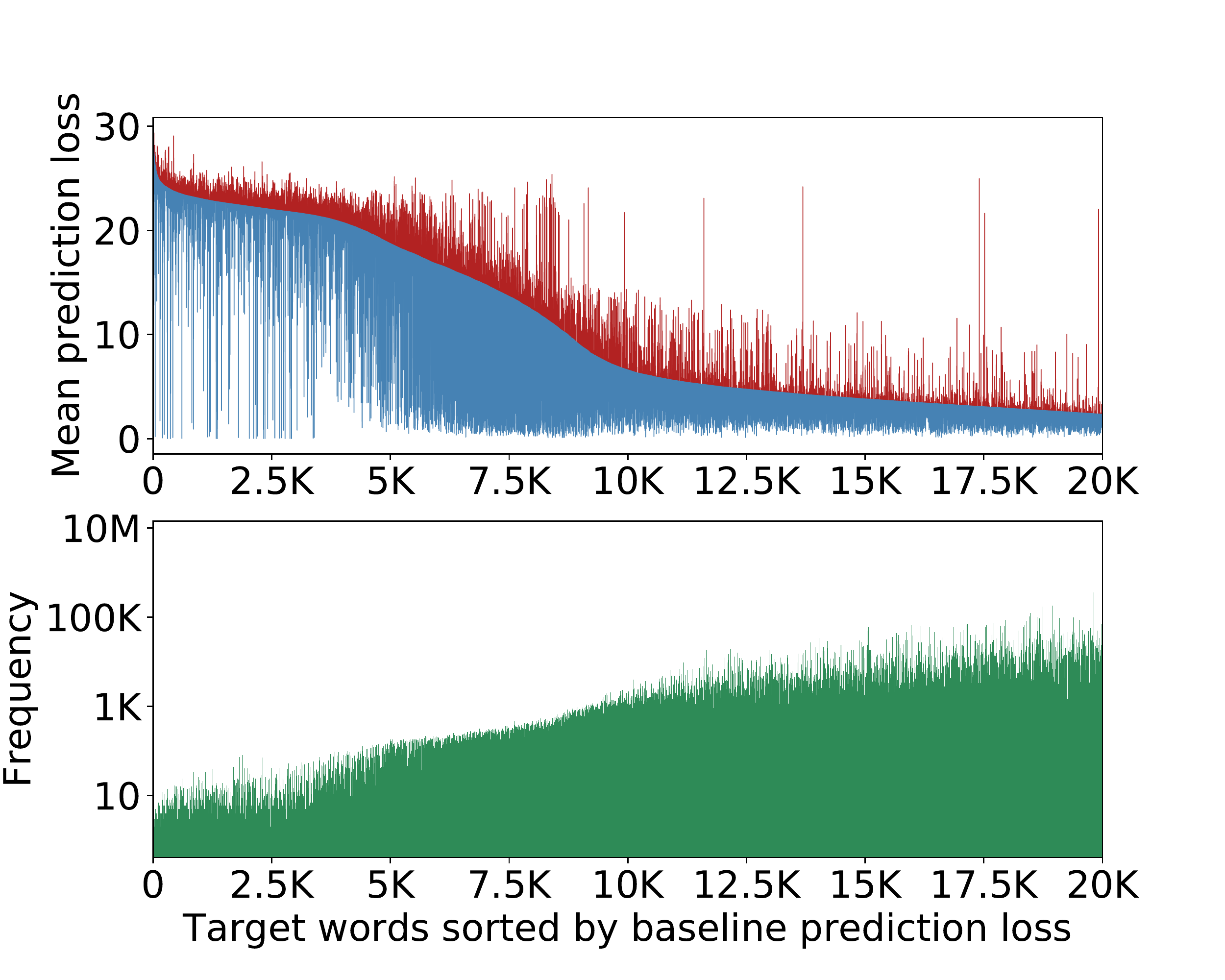}
\caption{Top: Changes in mean token prediction loss after re-training with synthetic data sorted by mean prediction loss of the baseline system. Decreases and increases in values are marked blue and red, respectively. Bottom: Frequencies (log) of target tokens in the baseline training data.}
\label{loss}
\end{figure}

As mentioned in Section~\ref{nmtsec}, the objective function of training an NMT system is to minimize $\mathcal{L}$ by minimizing the prediction loss $-\log p(y_t | y_{<t}, \vt{s}_n)$ for each target token in the training data.
The addition of monolingual data in the target language improves the estimation of the probability $p(Y)$ and consequently the model generates more fluent sentences.

\newcite{sennrich-haddow-birch:2016:P16-11} show that by using back-translation, the system with target-side monolingual data reaches a lower perplexity on the development set.
This is expected since the domain of the monolingual data is similar to the domain of the development set.
To investigate the model's accuracy independent from the domains of the data, we collect statistics of the target token prediction loss during training.

Figure~\ref{loss} shows changes of token prediction loss 
when training converges and the weights are verging on being stable.
The values are sorted by mean token prediction loss of the system trained on real parallel data.

We observe an effect similar to distributional smoothing \cite{chen1996empirical}: While prediction loss increases slightly for most tokens, the largest decrease in loss occurs for tokens with high prediction loss values.
This indicates that by randomly sampling sentences for back-translation, the model improves its estimation of tokens that were originally more difficult to predict, i.e., tokens that had a high prediction loss. 
%
Note that we compute the token prediction loss in just one pass over the training corpus with the final model and as a result it is not biased towards the order of the data.

This finding motivates us to further explore sampling criteria for back-translation that contribute considerably to the parameter estimation of the translation model. 
We propose that by oversampling sentences containing difficult-to-predict tokens we can maximize the impact of using the monolingual data.
After translating sentences containing such tokens and including them in the training data, the model becomes more robust in predicting these tokens.

In the next two sections, we propose several methods of using the target token prediction loss to identify the most rewarding sentences for back-translating and re-training the translation model. 

\section{Targeted Sampling for Difficult Words}

One of the main outcomes of using synthetic data is better estimation of words that were originally difficult to predict as measured by their high prediction losses during training.
In this section, we propose three variations of how to identify these words and perform sampling to target these words.

\begin{algorithm}
\small
\caption{Sampling for difficult words}\label{alg1}
 \hspace*{\algorithmicindent} {\textbf{Input:}} Difficult tokens \textit{$\mathfrak{D}=\{y_i\}_{i=1}^{D}$}, monolingual \\
 \hspace*{\algorithmicindent} corpus \textit{$\mathbb{M}$}, number of required samples \textit{N}   \\
\hspace*{\algorithmicindent} {\textbf{Output:}} Sampled sentences $S=\{S_i\}_{i=1}^{N}$ where each\\
 \hspace*{\algorithmicindent}   sentence $S_i$ is sampled from $\mathbb{M}$
 \begin{algorithmic}[1]
\Procedure{\textsc{DiffSampling} ($\mathfrak{D}, \mathbb{M}, N$):}{}
\State Initialize $S=\{\}$ 
\Repeat
\State Sample $S_c$ from $\mathbb{M}$
\ForAll{tokens $y$ in $S_c$}
\If{$y \in \mathfrak{D}$} 
	\State Add $S_c$ to $S$
\EndIf
\EndFor
\Until{$|S| = N$}
\State \textbf{return} $S$ 
\EndProcedure
\end{algorithmic}
\end{algorithm}

\begin{table*}[t!]
\begin{center}
\small%
\tabcolsep=2.5pt%
\begin{tabular}{@{ }l|ccccc|cccc@{}}
\toprule
   & \multicolumn{4}{c}{\textbf{De-En}} & & \multicolumn{4}{c}{\textbf{En-De}}  \\ \cmidrule{2-6} \cmidrule{7-10}
 \textbf{System}  &  test2014 & test2015 &  test2016 & test2017 & & test2014 &  test2015 &  test2016 &  test2017 \\ 
\midrule     
 \textsc{Baseline}$^\dagger$		& 26.7 &	27.6 &	32.5& 	28.1&& 21.2	&23.3&	28.0&	22.4\\
\textsc{Random}$^\dagger$ & 	28.7& 		29.7	&	36.3		&30.8&&	 24.0&	26.0&	30.7	&24.8\\ \midrule
\textsc{Freq} 	& 29.7&30.5	&37.5&31.4 &&24.2&27.0&31.7 &25.2\\			
\textsc{MeanPredLoss}$^\dagger$	 &	29.9 &		\textbf{30.9} 		&\textbf{37.8}	&\textbf{32.1}&	 &\textbf{24.7}&26.8&31.5&\textbf{25.5}\\ 
\textsc{MeanPredLoss + stdPredLoss} 	& 	\textbf{30.0}&		30.9		&37.7	&31.9 &&24.1&26.9  &31.0&25.3\\
\textsc{Preserve PredLoss ratio}	&		29.8	&30.9&37.4& 31.6& &24.5&\textbf{27.2}&\textbf{31.8} &25.5\\
\bottomrule
\end{tabular}
\end{center}
\caption{\label{meanresults}  German$\leftrightarrow$English translation quality (\textsc{Bleu}). Experiments marked $^\dagger$ are averaged over 3 runs. \textsc{MeanPredLoss} and \textsc{Freq} are difficulty criteria based on mean token prediction loss and token frequency respectively. \textsc{MeanPredLoss + stdPredLoss} is experiments favoring tokens with skewed prediction losses. \textsc{Preserve PredLoss ratio} preserves the  ratio of the distribution of difficult contexts.	}
\end{table*}

\subsection{Token Frequency as a Feature of Difficulty} 

Figure~\ref{loss} shows that the majority of tokens with high mean prediction losses have low frequencies in the training data.
Additionally, the majority of decreases in prediction loss after adding synthetic sentence pairs to the training data occurs with less frequent tokens.

Note that these tokens are not necessarily \textit{rare} in the traditional sense; in Figure~\ref{loss} the $10k$ less frequent tokens in the target vocabulary benefit most from back-translated data.

Sampling new contexts from monolingual data provides context diversity proportional to the token frequencies and less frequent tokens benefit most from new contexts.
Algorithm~\ref{alg1} presents this approach where the list of difficult tokens is defined as: 
\begin{align*}
\mathfrak{D} = \{\forall y_i \in V_t \colon freq(y_i) < \eta \}
\end{align*}
$V_t$ is the target vocabulary and $\eta$ is the frequency threshold for deciding on the difficulty of the token.

\subsection{Difficult Words with High Mean Prediction Losses}

In this approach, we use the mean losses to identify difficult-to-predict tokens. 
The mean prediction loss $\hat{\ell}(y)$ of token $y$ during training is defined as follows:
\begin{align*}\label{ave}
\hat{\ell}(y) = \frac{1}{n_y}\sum_{n=1}^{N}\sum_{t=1}^{|Y^{n}|} -\log p(y^{n}_t | y^{n}_{<t}, \vt{s}_n) \delta(y^n_t,y)
\end{align*}
where $n_y$ is the number of times token $y$ is observed during training, i.e., the token frequency of $y$, $N$ is the
number of sentences in the training data, $|Y^{n}|$ is the length of target sentence $n$, and $\delta(y^n_t,y)$ is the Kronecker delta function, which is 1 if $y^n_t = y$ and 0 otherwise.

By specifically providing more sentences for difficult words, we improve the model's estimation and decrease the model's uncertainty in prediction.
During sampling from the monolingual data, we select sentences that contain difficult words.

Algorithm~\ref{alg1} presents this approach where the list of difficult tokens is defined as: 
\begin{align*}
\mathfrak{D} = \{\forall y_i \in V_t \colon \hat{\ell}(y_i) > \mu \}
\end{align*}

$V_t$ is the vocabulary of the target language and $\mu$ is the threshold on the difficulty of the token.

\subsection{Difficult Words with Skewed Prediction Losses}

By using the mean loss for target tokens as defined above, we do not discriminate between differences in prediction loss for occurrences in different contexts. This lack of discrimination can be problematic for tokens with high loss variations. For instance, there can be a token with ten occurrences, out of which two have high and eight have low prediction loss values. 

We hypothesize that if a particular token is easier to predict in some contexts and harder in others, the sampling strategy should be context sensitive, allowing to target specific contexts in which a token has a high prediction loss. 
In order to distinguish between tokens with a skewed and tokens with a more uniform prediction loss distribution, we use both mean and standard deviation of token prediction losses to identify difficult tokens.

Algorithm~\ref{alg1} formalizes this approach where the list of the difficult tokens is defined as: 
\begin{align*}
\mathfrak{D} = \{\forall y_i \in V_t \colon \hat{\ell}(y_i) > \mu \land \sigma(\ell(y_i)) > \rho \}
\end{align*}

$\hat{\ell}(y_i)$ is the mean and $\sigma(\ell(y_i))$ is the standard deviation of prediction loss of token $y_i$, $V_t$ is the vocabulary list of the target language, and $\mu$ and $\rho$ are the thresholds for deciding on the difficulty of the token.

 \begin{algorithm}
\small
\caption{Sampling with ratio preservation}\label{alg2}
 \hspace*{\algorithmicindent} \textbf{Input:} Difficult tokens and the corresponding sentences \\
  \hspace*{\algorithmicindent} in the bitext \textit{$\mathfrak{D}=\{y_t, Y_{y_t}=[y_1, \ldots, y_t, \ldots, y_m]\}$}, \\
   \hspace*{\algorithmicindent} monolingual corpus \textit{$\mathbb{M}$}, number of required samples \textit{N} \\
\hspace*{\algorithmicindent} \textbf{Output:} Sampled sentences $S=\{S_i\}_{i=1}^{N}$ where each\\
 \hspace*{\algorithmicindent} sentence $S_i$ is sampled from $\mathbb{M}$
 \begin{algorithmic}[1]
\Procedure{\textsc{PredLossRatioSampling}($\mathfrak{D}, \mathbb{M}, N$): }{}
\State Initialize $S=\{\}$
\State $H(y_t) = \frac{ N \times \mid(y_t, \boldsymbol{\cdot})\in \mathfrak{D}\mid}{\mid(y_{\boldsymbol{\cdot}}, \boldsymbol{\cdot})\in \mathfrak{D}\mid}$
\Repeat
\State Sample $S_c$ from $\mathbb{M}$
\ForAll{tokens $y$ in $S_c$}
\If{$|y \in S| < H(y)$} 
	\State Add $S_c$ to $S$
\EndIf
\EndFor
\Until{$|S| = N$}
\State \textbf{return} $S$ 
\EndProcedure
\end{algorithmic}
\end{algorithm}

 \subsection{Preserving Sampling Ratio of Difficult Occurrences}

Above we examined the mean of prediction loss for each token over all occurrences, in order to identify difficult-to-predict tokens.
However, the uncertainty of the model in predicting a difficult token varies for different occurrences of the token: one token can be easy to predict in one context, and hard in another.
While the sampling step in the previous approaches targets these tokens, it does not ensure that the distribution of sampled sentences is similar to the distribution of problematic tokens in difficult contexts.

To address this issue, we propose an approach where we target the number of times a token occurs in difficult-to-predict contexts and sample sentences accordingly, thereby ensuring the same ratio as the distribution of difficult contexts.
If token $y_1$ is difficult to predict in two contexts and token $y_2$ is difficult to predict in four contexts, the number of sampled sentences containing $y_2$ is double the number of sampled sentences containing $y_1$.
Algorithm~\ref{alg2} formalizes this approach.
 
 \subsection{Results}
 
We measure the translation quality of various models for German$\rightarrow$English and English$\rightarrow$German translation tasks.
The results of the translation experiments are presented in Table~\ref{meanresults}.
As baseline we compare our approach to \newcite{sennrich-haddow-birch:2016:P16-11}. 
For all experiments we sample and back-translate sentences from the monolingual data, keeping a one-to-one ratio of back-translated versus original data (\mbox{1:1}).

We set the hyperparameters $\mu$, $\rho$, and $\eta$ to 5, 10, and 5000 respectively.
The values of the hyperparameters are chosen on a small sample of the parallel data based on the token loss distribution. 

As expected using random sampling for back-translation improves the translation quality over the baseline.
However, all targeted sampling variants in turn outperform random sampling.
Specifically, the best performing model for German$\rightarrow$English, \textsc{MeanPredLoss}, uses the mean of prediction loss for the target vocabulary to oversample sentences including these tokens.

For the English$\rightarrow$German experiments we obtain the best translation performance when we preserve the prediction loss ratio during sampling.

We also observe that even though the model targeting tokens with skewed prediction loss distributions ($\textsc{MeanPredLoss + stdPredLoss}$) improves over random selection of sentences, it does not outperform the model using only mean prediction losses. 

\section{Context-Aware Targeted Sampling}

In the previous section, we proposed methods for identifying difficult-to-predict tokens and performed targeted sampling from monolingual data.
While the objective was to increase the occurrences of difficult tokens, we ignored the context of these tokens in the sampled sentences. 

Arguably, if a word is difficult to predict in a given context, providing more examples of the same or similar context can aid the learning process.
In this section, we focus on the context of difficult-to-predict words and aim to sample sentences that are similar to the corresponding difficult context.

The general algorithm is described in Algorithm~\ref{alg3}.
In the following sections, we discuss different definitions of the local context (line~\ref{cont1} and line~\ref{cont2}) and similarity measures (line~\ref{simss}) in this algorithm, and report the results.

\subsection{Definition of Local Context}

Prediction loss is a function of the source sentence and the target context.
One reason that a token has high prediction loss is that the occurrence of the word is a deviation from what occurs more frequently in other occurrences of the context in the training data.
This indicates an infrequent event, in particular a rare sense of the word, a domain that is different from other occurrences of the word, or an idiomatic expression. 

\begin{table}[htb!]
\begin{center}\small
\begin{tabularx}{\columnwidth}{lX}
\toprule
\textit{source} & wer glaube, dass das ende, sobald sie in Deutschland ank$\mid${\"a}$\mid$men, ir$\mid$re, erz{\"a}hlt \textbf{B$\mid$ahr}. \\
\textit{reference } & if you think that this stops as soon as they arrive in Germany, you'd be wrong, says \textbf{B$\mid$ahr}.\\
\textit{NMT output} & who believe that the end, as soon as they go to Germany, tells \textbf{B$\mid$risk}.\\
\bottomrule
\end{tabularx}
\end{center}
\caption{\label{exbpe1} An example from the synthetic data where the word \textit{B$\mid$ahr} is incorrectly translated to \textit{B$\mid$risk}. Subword unit boundaries are marked with `$\mid$'.}
\end{table}

We identify \textit{pairs} of tokens and sentences from parallel data where in each pair the NMT model assigns high prediction loss to the token in the given context. 
Note that a token can occur several times in this list, since it can be considered as difficult-to-predict in different sentences.

We propose two approaches to define the local context of the difficult token:

\paragraph{Neighboring tokens}

A straightforward way is to use positional context: tokens that precede and follow the target token, typically in a window of $w$ tokens to each side.
For sentence $S$ containing a difficult token at index $i$ the \textit{context} function in Algorithm~\ref{alg3} is:
\begin{align*}
context(S, i) = [S^{i-w}, \ldots, S^{i-1}, S^{i+1}, \ldots, S^{i+w}]
\end{align*}

where $S^{j}$ is the token at index $j$ in sentence $S$.

\begin{table}[htb!]
\begin{center}\small
\begin{tabularx}{\columnwidth}{X}
\toprule
 \textit{Sentence from bitext containing difficult token:}\\ \midrule
He attended \textbf{Stan}$\mid$ford University, where he double maj$\mid$ored in Spanish and History.  \\  
  \midrule
  \textit{Sampled sentences from monolingual data:} \\ 
  \midrule
   $-$ The group is headed by Aar$\mid$on K$\mid$ush$\mid$ner, a \textbf{Stan}$\mid$ford University gradu$\mid$ate who formerly headed a gre$\mid$eting card company. \\ 
   $-$ Ford just opened a new R\&D center near \textbf{Stan}$\mid$ford University, a hot$\mid$bed of such technological research. \\
   $-$ Joe Grund$\mid$fest, a professor and a colleague at \textbf{Stan}$\mid$ford Law School, outlines four reasons why the path to the IP$\mid$O has become so steep for asp$\mid$iring companies. \\
\bottomrule
\end{tabularx}
\end{center}
\caption{\label{exbpe2} Results of targeted sampling for the difficult subword unit \textit{`Stan'}.}
\end{table}

\paragraph{Neighboring subword units}

In our analysis of prediction loss during training, we observe that several tokens that are difficult to predict are indeed subword units.
Current state-of-the-art NMT systems apply BPE to the training data to address large vocabulary challenges \cite{sennrich-haddow-birch:2016:P16-12}.

By using BPE the model generalizes common subword units towards what is more frequent in the training data.
This is inherently useful since it allows for better learning of less frequent words.
However, a side effect of this approach is that at times the model generates subword units that are not linked to any words in the source sentence.

As an example, in Table~\ref{exbpe1} German source and English reference translation show this problem.
The word \textit{B$\mid$ahr} consisting of two subword units is incorrectly translated into \textit{B$\mid$risk}.

We address the insufficiency of the context for subword units with high prediction losses by targeting these tokens in sentence sampling.

Algorithm~\ref{alg3} formalizes this approach in sampling sentences from the monolingual data.
For a sentence $S$ containing a difficult subword at index $i$, the context function is defined as:
\begin{align*}
context(S, i) = [ \ldots, S^{i-1}, S^{i+1}, \ldots ]
\end{align*}
where every token $S^{j}$ in the local context is a subword unit and part of the same word as $S^i$.
Table~\ref{exbpe2} presents examples of sampled sentences for the difficult subword unit \textit{`Stan'}.

\begin{table}[htb!]
\begin{center}\small
\begin{tabularx}{\columnwidth}{X}
\toprule
 \textit{Sentence from bitext containing difficult word:}\\ \midrule
  Bud$\mid$dy Hol$\mid$ly was part of the first group induc$\mid$ted into the \textbf{Rock} and R$\mid$oll Hall of F$\mid$ame on its formation in 1986.  \\  \midrule
  \textit{Sampled sentences from monolingual data:} \\ \midrule
 $-$ A 2008 \textbf{Rock} and R$\mid$oll Hall of F$\mid$ame induc$\mid$t$\mid$ee, Mad$\mid$onna is ran$\mid$ked by the Gu$\mid$inn$\mid$ess Book of World Rec$\mid$ords as the top-selling recording artist of all time.\\
  $-$ The \textbf{Rock} and R$\mid$oll Hall of Fam$\mid$ers gave birth to the California rock sound. \\
  $-$ The winners were chosen by 500 voters, mostly musicians and other music industry veter$\mid$ans, who belong to the \textbf{Rock} and R$\mid$oll Hall of F$\mid$ame Foundation.\\
\bottomrule
\end{tabularx}
\end{center}
\caption{\label{context} Results of context-aware targeted sampling for the difficult token \textit{`Rock'} }
\end{table}

\begin{algorithm}
\small
\caption{Sampling with context}\label{alg3}
  \hspace*{\algorithmicindent} \textbf{Input:} Difficult tokens and the corresponding sentences \\
  \hspace*{\algorithmicindent} in the bitext \textit{$\mathfrak{D}=\{y_t, Y_{y_t}=[y_1, \ldots, y_t, \ldots, y_m]\}$}, \\
   \hspace*{\algorithmicindent} monolingual corpus \textit{$\mathbb{M}$}, context function \textit{$context$}, \\ 
   \hspace*{\algorithmicindent} number of required samples \textit{N}, similarity threshold $s$ \\
\hspace*{\algorithmicindent} \textbf{Output:} Sampled sentences $S=\{S_i\}_{i=1}^{N}$ where each \\   
 \hspace*{\algorithmicindent} sentence $S_i$ is sampled from $\mathbb{M}$
 \begin{algorithmic}[1]
\Procedure{\textsc{CntxtSampling}($\mathfrak{D}, \mathbb{M}, context, N, s$): }{}
\State Initialize $S=\{\}$
\Repeat
\State Sample $S_c$ from $\mathbb{M}$
\ForAll{tokens $y_t$ in $S_c$}
	\If{$y_t \in \mathfrak{D}$} 
	\State $C_m \leftarrow context(S_c,$ indxof$(S_c, y_t))$ \label{cont1}
	\ForAll{$Y_{y_t}$}
		\State $C_p \leftarrow context(Y_{y_t},$  indxof$(Y_{y_t}, y_t)) $ \label{cont2}
		\If{Sim$(C_m, C_p) > s$}  \label{simss}
	 		\State Add $S_c$ to $S$
		\EndIf
	\EndFor
	\EndIf
\EndFor
\Until{$|S| = N$}
\State \textbf{return} $S$ 
\EndProcedure
\end{algorithmic}
\end{algorithm}

\begin{table*}[t!]
\begin{center}
\small%
\tabcolsep=2.5pt%
\begin{tabular}{@{ }lll|ccccc|ccccc@{}}
\toprule
 &&& \multicolumn{4}{c}{\textbf{De-En}} & & \multicolumn{4}{c}{\textbf{En-De}}  \\ \cmidrule{4-8} \cmidrule{9-12}
 \textbf{System}  & &&  test2014 & test2015 &  test2016 & test2017 & & test2014 &  test2015 &  test2016 &  test2017 \\ 
 \midrule    
 \textsc{Baseline} $^\dagger$ & && 26.7 &	27.6 &	32.5	&28.1&& 21.2	&23.3&	28.0&	22.4\\
\textsc{Random} $^\dagger$ & &&  28.7& 		29.7	&	36.3	&	30.8&&	 24.0&	26.0&	30.7	&24.8\\ \midrule
 \textbf{Difficulty criterion} & \textbf{Context} & \textbf{Similarity}       \\ \midrule
\textsc{Freq} &  \textsc{tokens} &\textsc{emb}  &	 30.0&30.8&37.6&31.7& &24.4 &   26.3 &31.5&25.6 \\	
\textsc{PredLoss} & \textsc{Swords}  & \textsc{match}		& 29.1&30.1&	36.9&31.0&&23.8&26.2 &28.8&23.2\\			
\textsc{PredLoss} & \textsc{tokens} & \textsc{match} 	&	29.7 & 	30.6	&37.6&	31.8 & &24.3  &27.4&31.6 & 25.5\\ 
\textsc{PredLoss} & \textsc{tokens}  & \textsc{emb }	 &29.9&30.8	&37.7&31.9&  &24.5  &\textbf{27.5}&31.7 & 25.6\\
\textsc{PredLoss} &\textsc{sentence}  & \textsc{emb }		& 24.9&	25.5	&30.1&	26.2 &  &22.0&	24.6&	27.9	&22.5\\				
\textsc{MeanPredLoss } & \textsc{tokens}  & \textsc{emb } &   \textbf{30.2}	&\textbf{31.4}	&\textbf{37.9}	&\textbf{32.2}&&\textbf{24.4}& 27.2&\textbf{31.8}&\textbf{25.6}\\
\bottomrule
\end{tabular}
\end{center}
\caption{\label{contextresults} German$\leftrightarrow$English translation quality (\textsc{Bleu}). Experiments marked $^\dagger$ are averaged over 3 runs. \textsc{PredLoss} is the contextual prediction loss and \textsc{MeanPredLoss} is the average loss. \textsc{token} and \textsc{Sword} are context selection definitions from neighboring tokens and subword units respectively. Note that token includes both subword units and full words. \textsc{emb} is computing context similarities with token embeddings and \textsc{match} is comparing the context tokens.  }
\end{table*}

\subsection{Similarity of the Local Contexts}

In context-aware targeted sampling, we compare the context of a sampled sentence and the difficult context in the parallel data and select the sentence if they are \textit{similar}.
In the following, we propose two approaches for quantifying the similarities. 

\paragraph{Matching the local context}

In this approach we aim to sample sentences containing the difficult token, matching the exact context to the problematic context.
By sampling sentences that match in a local window with the problematic context and differ in the rest of the sentence, we have more instances of the difficult token for training.

Algorithm~\ref{alg3} formalizes this approach where the similarity function is defined as:
\begin{align*}
\text{Sim}(C_m, C_p) = \frac{1}{c} \sum_{i=1}^{c}\delta(C_m^i, C_p^i)
\end{align*}

$C_m$ and $C_p$ are the contexts of the sentences from monolingual and parallel data respectively, and $c$ is the number of tokens in the contexts.

\paragraph{Word representations}

Another approach to sampling sentences that are similar to the problematic context is to weaken the matching assumption.
Acquiring sentences that are similar in subject and not match the exact context words allows for lexical diversity in the training data.
We use embeddings obtained by training the Skipgram model \cite{mikolov2013efficient} on monolingual data to calculate the similarity of the two contexts.

For this approach we define the similarity function in Algorithm~\ref{alg3} as:
\begin{align*}
\text{Sim}(C_m, C_p)  = cos(\vt{v}(C_m), \vt{v}(C_p))
\end{align*}
where $\vt{v}(C_m)$ and $\vt{v}(C_p)$ are the averaged embeddings of the tokens in the contexts.

Table~\ref{context} presents examples of sampled sentences for the difficult word \textit{rock}.
In this example, the context where the word \textit{`Rock'} has high prediction loss is about the \textit{music genre} and not the most prominent sense of the word, \textit{stone}.
Sampling sentences that contain this word in this particular context provides an additional signal for the translation model to improve parameter estimation.

 \subsection{Results}

The results of the translation experiments are given in Table~\ref{contextresults}.
In these experiments, we set the hyperparameters $s$ and $w$ to 0.75 and 4, respectively.
Comparing the experiments with different similarity measures, \textsc{match} and \textsc{emb}, we observe that in all test sets we achieve the best results when using word embeddings.
This indicate that for targeted sampling, it is more beneficial to have diversity in the context of difficult words as opposed to having the exact ngrams. 

When using embeddings as the similarity measure, it is worth noting that with a context of size 4 the model performs very well but fails when we increase the window size to include the whole sentence. 

The experiments focusing on subword units (\textsc{sword}) achieve improvements over the baselines, however they perform slightly worse than the experiments targeting tokens (\textsc{token}).

The best \textsc{Bleu} scores are obtained with the mean of prediction loss as difficulty criterion (\textsc{meanPredLoss}) and using word representations to identify the most similar contexts.
We observe that summarizing the distribution of the prediction losses by its mean is more beneficial than using individual losses.
Our results motivate further explorations of using context for targeted sampling sentences for back-translation.

\section{Conclusion}

In this paper we investigated the effective method of back-translation for NMT and explored alternatives to select the monolingual data in the target language that is to be back-translated into the source language to improve translation quality.

Our findings showed that words with high prediction losses in the target language benefit most from additional back-translated data.

As an alternative to random sampling, we proposed targeted sampling and specifically targeted words that are difficult to predict.
Moreover, we used the contexts of the difficult words by incorporating context similarities as a feature to sample sentences for back-translation.
We discovered that using the prediction loss to identify weaknesses of the translation model and providing additional synthetic data targeting these shortcomings improved the translation quality in German$\leftrightarrow$English by up to 1.7 \textsc{Bleu} points.

\section*{Acknowledgments}
This research was funded in part by the Netherlands Organization for Scientific Research (NWO) under project numbers 639.022.213 and 612.001.218.
 We also thank NVIDIA for their hardware support and the anonymous reviewers for their helpful comments.
\bibliography{acl2018}
\bibliographystyle{acl_natbib_nourl}

\end{document}